\def\year{2022}\relax
\documentclass[letterpaper]{article} 
\usepackage{aaai22}  
\usepackage{times}  
\usepackage{helvet}  
\usepackage{courier}  
\usepackage[hyphens]{url}  
\usepackage{graphicx} 
\usepackage{booktabs}
\usepackage{natbib}  
\usepackage{caption} 
\DeclareCaptionStyle{ruled}{labelfont=normalfont,labelsep=colon,strut=off} 
\usepackage[switch]{lineno}

\usepackage{algorithm}
\usepackage{algorithmic}
\usepackage{amsmath}  
\usepackage{multirow}
\usepackage{amssymb}
\usepackage{newfloat}
\usepackage{listings}
\floatstyle{ruled}
\newfloat{listing}{tb}{lst}{}
\floatname{listing}{Listing}
\author{
  Anonymous AAAI 2021 Submission \\
  Paper ID: 592\\ 
}

\title{Dual Contrastive Learning for General Face Forgery Detection}

\author {
    Ke Sun\textsuperscript{\rm 1},
    Taiping Yao\textsuperscript{\rm 2},
    Shen Chen\textsuperscript{\rm 2},
    Shouhong Ding\textsuperscript{\rm 2}\footnote{Corresponding authors.},
    Jilin Li \textsuperscript{\rm 2}, 
    Rongrong Ji\textsuperscript{\rm 1}
    \footnotemark[\value{footnote}]
    \\ 
}
\affiliations {
    \textsuperscript{\rm 1}Media Analytics and Computing Lab, Department of Artificial Intelligence, \\School of Informatics, Xiamen University, 361005, China \\
    \textsuperscript{\rm 2}Youtu Lab, Tencent, China\\
    skjack@stu.xmu.edu.cn, rrji@xmu.edu.cn\\
    \{taipingyao, kobeschen, ericshding, jerolinli\}@tecent.com
    
}

\usepackage{bibentry}

\begin{document}

\maketitle

\begin{abstract}
With various facial manipulation techniques arising, face forgery detection has drawn growing attention due to security concerns. Previous works always formulate face forgery detection as a classification problem based on cross-entropy loss, which emphasizes category-level differences rather than the essential discrepancies between real and fake faces, limiting model generalization in unseen domains. 
To address this issue, we propose a novel face forgery detection framework, named Dual Contrastive Learning (DCL), which specially constructs positive and negative paired data and performs designed contrastive learning at different granularities to learn generalized feature representation. 
Concretely, combined with the hard sample selection strategy, Inter-Instance Contrastive Learning (Inter-ICL) is first proposed to promote task-related discriminative features learning by especially constructing instance pairs. Moreover, to further explore the essential discrepancies, Intra-Instance Contrastive Learning (Intra-ICL) is introduced to focus on the local content inconsistencies prevalent in the forged faces by constructing local-region pairs inside instances. 
Extensive experiments and visualizations  on several datasets demonstrate the generalization 
of our method against the state-of-the-art competitors.

\end{abstract}

\section{Introduction}
Over the past few years, face forgery methods have achieved significant success and received lots of attention in the computer vision community~\cite{thies2015real,rossler2019faceforensics++,dolhansky2020deepfake,wang2020face,gu2021spatiotemporal}. As such techniques can generate high-quality fake videos that are even indistinguishable for human eyes, they can easily be abused by malicious users to cause severe societal problems or political threats. To mitigate such risks, it is of paramount importance to develop effective methods for detecting face forgery.

Early works~\cite{afchar2018mesonet,stehouwer2019detection,dolhansky2020deepfake} treat face forgery detection as a binary classification problem and use the convolutional neural network (CNN) to distinguish the authenticity of the face. These methods achieve considerable performance in the intra-domain scenario, where the training and test sets manifest similar data distributions. However, faces in real applications are varied from that in training set in terms of the camera type, pre-processing, compression rate, and attack method, \textit{e.t.c}. These unseen domain gaps bring severe performance drops, thus limiting broader applications.

Recently, several works~\cite{li2020face,sun2021domain,liu2021spatial} are devoted to \textit{general face forgery detection}~\cite{sun2021domain} to relieve the generalizing problem. Face X-ray~\cite{li2020face} designs supervision named face X-ray, which focuses on the blending boundary induced by the image blending process. LTW~\cite{sun2021domain} weights training samples via meta-learning. Besides, some attempt to obtain information from frequency domains, such as DCT~\cite{qian2020thinking}, spatial-phase~\cite{liu2021spatial} and SRM~\cite{luo2021generalizing}. However, these methods can be attributed to a binary classification network based on cross-entropy loss, which we argue is not suitable for general face forgery detection. Specifically, traditional cross-entropy based methods assume that all instances within each category should be close in feature distribution while ignoring the unique information of each sample. This information is proven in ~\cite{zhao2020makes} that could provide transferability knowledge for unseen forgery faces. Without additional constraints, a common cross-entropy classification framework is prone to overfitting on specific forged patterns~\cite{luo2021generalizing}. Moreover, the quality of forgery faces is different, and optimizing each sample equally as in the traditional framework makes it difficult to uncover the underlying forgery clues and is not conducive to generalization~\cite{liu2021dual,sun2021domain}. Therefore, a new framework is urgently needed to address the above issues.

In this paper, motivated by contrastive learning~\cite{he2020momentum} that has been shown to outperform its supervised counterpart in transfer learning~\cite{zhao2020makes}, we propose a Dual Contrast Learning (DCL) framework for general facial forgery detection, as shown in Fig.~\ref{fig:main}.
Since DCL requires two samples from different views of the same image as inputs, we first generate different data views as positive pairs via specially designed data transformations which can eliminate the task-independent information, such as background and face structure, \textit{e.t.c}.
Second, to preserve the instance transferability, we design the Inter-instance Contrastive Learning (Inter-ICL) to mine the association between instances. Specifically, Inter-ICL pulls the positive pairs closer together while pulling the true negatives of each training instance away. The variance among instances can be preserved by alignment and uniformity~\cite{wang2020understanding}. We also propose a new hard sample selection strategy by comparing samples with their negative prototypes. This strategy can provide effective gradients to the contrastive loss and highlight samples with essential forgery clues. 
Third, considering the local inconsistencies prevalent in the forged faces, we design Intra-instance Contrastive Learning (Intra-ICL), which contrasts fake and real parts within forgery face to mine the general essential forgery clues.

\begin{itemize}
    \item We propose a novel Dual Contrastive Learning (DCL) for general face forgery detection, which specially  constructs  positive  and  negative  data pairs  and  performs  contrastive  learning  at  different granularities to further improve the generalization.
    \item 
    We specially design Inter-Instance Contrastive Learning and Intra-Instance Contrastive Learning based on instance pairs among samples and local-region pairs within samples respectively to learn task-related essential features.
    \item Extensive experiments and visualizations demonstrate the effectiveness of our method against the state-of-the-art competitors.
\end{itemize}
\section{Related Work}
\subsection{Face Forgery Detection}
Face forgery detection is a classical problem in computer vision and graphics. 
Earlier studies focus on hand-crafted features such as eye blinking~\cite{Li2018InIO}, inconsistent head poses~\cite{yang2019exposing} and visual artifacts~\cite{matern2019exploiting}.
With the tremendous success of deep learning, convolutional neural network (CNN) is wildely used to face forgery detection task and achieved better performance.
For example, ~\citet{stehouwer2019detection,zhao2021multi} highlighted the manipulated regions via attention mechanism.
\citet{frank2020leveraging} first discovered the difference between real and forgery face under frequency domain. 
Subsequently, many works~\cite{qian2020thinking,masi2020two,chen2021local,liu2021spatial} leverage frequency clues as the supplement to RGB information. Although the aforementioned methods achieve promising results in intra-domain where the data distributions in training set and test set are the same, the performance drops significantly when facing the unseen domain scenario.

Recently, some work focusing on general face forgery detection has been proposed. Face X-ray~\cite{li2020face} is supervised by the forged boundary that widely existed in blending operation. LTW~\cite{sun2021domain} weight samples and provide gradient regularization via meta-learning. ~\citet{luo2021generalizing} depress the texture bias via SRM operation to avoid overfitting on image content.
However, these methods inherited from image classification models emphasize category-level differences rather than the essential discrepancies between real and fake images.
To tackle these issues, we introduce dual-granularity contrastive learning framework to control the intra-class variance and preserve the transferability.


\subsection{Supervised Contrastive Learning}
Contrastive learning ~\cite{he2020momentum} has achieved great success in self-supervised representation. \citet{khosla2020supervised} extend it to the fully-supervised setting that can effectively leverage label information. The advantage of supervised contrastive learning (SCL) can be attributed twofold:
Firstly, it can be used to change the feature distributions. Based on SCL, ~\citet{bukchin2021fine} reduce the intra-class variance and eliminate the intra-class distinctions between sub-classes in the coarse-to-fine few-shot task, while ~\citet{wang2021exploring} enforce pixel embeddings belonging to the same semantic class to be more similar than embeddings from different classes in semantic segmentation task.
Secondly, SCL can eliminate the task-independent information by constructing positive data pairs. For example, \citet{lo2021clcc} decouple the correlation of the image scenes and illuminant via supervised contrastive learning for color constancy. ~\citet{wang2021exploring}, focus on the 3D face presentation attack detection,
learns discriminative features by comparing image pairs with diverse contexts. 
For face forgery detection, we leverage specially designed data transform methods as positive pairs and further boost the unseen domain performance via intra-instance contrastive learning and inter-instance contrastive learning.

\section{Proposed Method}
\begin{figure*}[h]
    \begin{center}
    \includegraphics[width=0.9\textwidth]{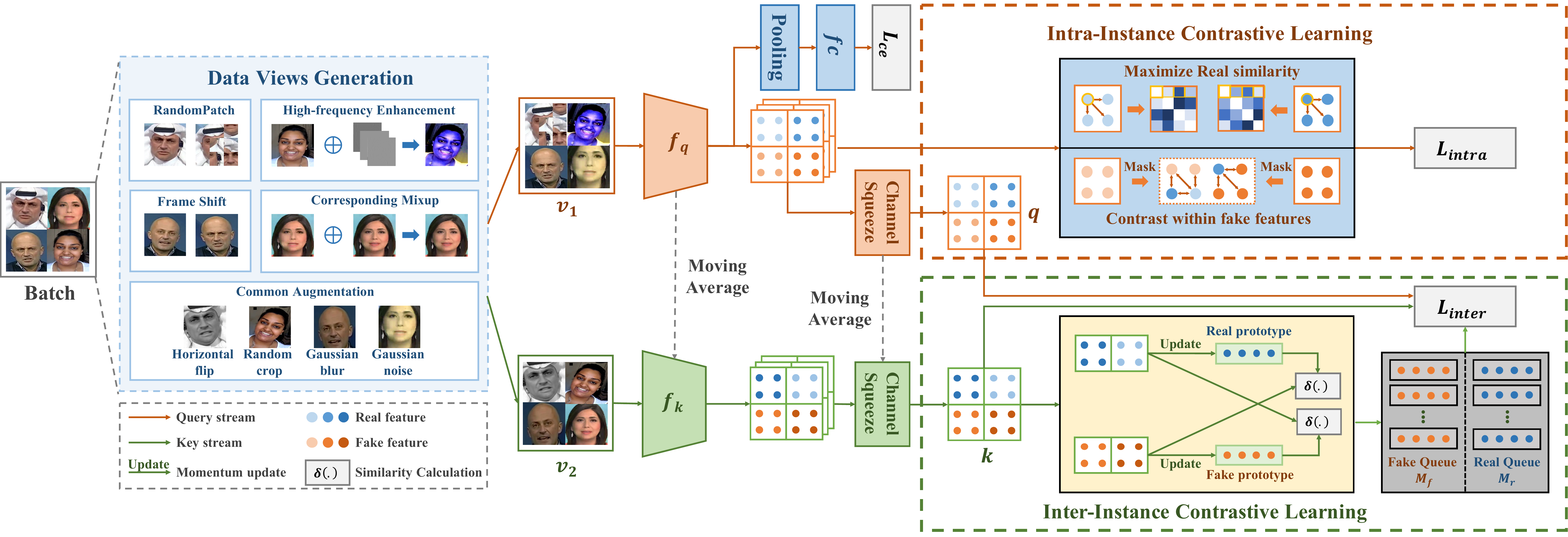}
    \end{center}
       \caption{Overview of our proposed DCL framework. Given training images, we first transform them into two different views via the Data Views Generation module. Then the Intra-instance contrastive learning module and Inter-instance contrastive learning module are proposed to learn general features.
       }
    \label{fig:main}
\end{figure*}
In this section, we introduce our Dual Contrastive Learning (DCL) framework for general face forgery detection, which simultaneous contrast features between different instances and within the instance.
As shown in Fig~\ref{fig:main}, DCL trains the model via contrastive learning framework in a supervised manner. The beginning of the DCL is Data Views Generation (DVG) module to generate different views of inputs by special designed data augmentation, then features are extracted from the well-designed supervised contrastive learning architecture. Subsequently, the Inter-Instance Contrastive Learning module and Intra-Instance Contrastive Leaning module are used to arrange the feature distribution and enhance the inconsistency of forgery faces, respectively.

\subsection{Data Views Generation}
\label{section:1}
In the contrastive learning framework~\cite{he2020momentum}, features are pulled closer if they are encoded views of the same image. Thus, it is important to generate different views as the positive pair.
Traditional contrastive learning uses common data augmentation such as horizontal flip, random crop and, gaussian blur to generate views. Different from the common classification task, the key of the general forgery face views' generalization is to eliminate task-irrelevant contextual factors such as high-level face content, specific manipulated texture, background information \textit{e.t.c}. Thus, we leverage the following specifically designed operations to generate different views:
\textbf{1) RandomPatch.} To destroy the structure of the input face and highlight the focus on the forgery clues, we divide the input face into $k \times k$ patches, and then randomly shuffle them.
\textbf{2) High-frequency enhancement.}
Existing works~\cite{fridrich2012rich} have proved that the high-frequency features can help to boost the generalization ability. Take inspiration from it, we combine the SRM feature with the original image to enhance the high-frequency information. 
\textbf{3) Frame shift.} To reduce the influence of different expressions and motions on forged extraction, we choose different frames of the same video as different views.
\textbf{4) Corresponding mixup.} In order to
eliminate some obviously forgery clues and mine the essential features, we apply a mixup operation between fake images and its corresponding real images. Note that this operation is only used when the input is a fake image.
All these methods are chosen with a certain probability and random combination becomes the final operations denotes as $v_1(.)$ and $v_2(.)$.

\subsection{Architecture of Contrastive Leaning}
\label{section:2}
The input data $x_i\in{R^{H{\times}W\times3}}$ with label $y_i \in{\{0,1\}}$ is firstly transformed into two different views $v_1(x_i)$ and $v_2(x_i)$ via data views generation module. 
Then two views are fed into CNN based query encoder $f_q$ and key encoder $f_k$ to obtain feature maps $f_q(v_1(x_i))\in R^{C \times H' \times W'}$ and $f_k(v_2(x_i))\in R^{C \times H' \times W'}$. 
Similar to the MoCo~\cite{he2020momentum}, the parameters of key encoder $\theta^{'}$ is updated via exponential moving-average strategy from query encoder parameters $\theta$:
\begin{equation}
    \theta^{'} = \beta \theta^{'} + (1-\beta)\theta,
    \label{equation:eq1}
\end{equation}
\vspace{-2pt}
where $\beta$ is exponential hyper-parameter. 
Since it is important for face forgery detection task to locate the forgery clues in spatial dimensional, unlike traditional contrastive learning, we use spatial-wise instead of channel-wise features as contrastive objectives. Specifically, $1\times1$ convolution operation is applied to squeeze the channel dimensional to get query $q \in R^{H'\times W'}$ and key $k\in R^{H'\times W'}$. 
Subsequently, to perform classification and make full use of label information, a fully connected classifier $fc$ is inserted after query feature extraction.
We formulated the binary cross-entropy loss $L_{ce}$ as follows:
\begin{equation}
    L_{ce} = y\log{y^{'}} + (1-y)\log(1-y^{'}),
    \label{equation:eq2}
\end{equation}
where $y^{'}$ is the final predicted probability and $y$ denoted the corresponding ground-truth label. During testing, only query encoder $f_{q}$ and classifier $fc$ is used to get final prepositions.

\subsection{Inter-Instance Contrastive Learning.}
\label{section:3}
To preserve the instance discrimination of unique sample, we design an inter-instance contrastive learning module which pulls close the embedding of the different views of the same image and pull away the true negative samples on the hyperspherical plane.

Specifically, we maintain two feature queues:  real queue $M_{r}$ and fake queue $M_{f}$ to construct the negative sample of the corresponding query.
The normalized cosine similarity is used as the metric of features  denoted as  $\delta(u,v) = \frac{u}{||u||}.\frac{v}{||v||} $.
Our inter-instance contrastive loss based upon the popular InfoNCE loss ~\cite{gutmann2010noise} can be written as: 

\begin{equation}
    L_{inter} = -\log\frac{e^{\delta(q,k) / \tau}}{e^{\delta(q,k)/ \tau} + \sum_{k_{m}\in K_{q}^{-}}e^{\delta(q,k_m)/ \tau}
    },
    \label{equation:eq3}
\end{equation}

where $\tau$ is a temperature parameter which controls the scale of distribution. $K_{q}^{-}$ represents the negative set of $q$, \textit{i.e.} $K_{q}^{-} = M_{f}$ when $q$ belongs to real, and $K_{q}^{-} = M_{r}$ when $q$ belongs to fake.
Unlike existing common supervised contrastive learning methods ~\cite{khosla2020supervised} which maximizes the invariance among the same category views, 
$L_{cl}$ only maximizes the invariance between two different views of single input, which decouples the impact of irrelevant forgery traces and does avoid the embedding instance of the same class in the proximity of one another. Thus, the variance of the intra-class distributions can be guaranteed so that more transferability knowledge is preserved. 

\noindent\textbf{Hard Sample Generation.}
\citet{robinson2020contrastive,wang2021exploring} found that the hard negative pair selection is crucial for contrastive learning.
Meanwhile, \citet{sun2021domain} shows high-quality samples are more conducive to improving the generalization of the model. 
Thus, to bring more gradient contributions for the $L_{inter}$  and mine the essential general forgery features, we design a novel hard sample generation strategy to generate our feature queues $M_{f}$ and $M_{r}$.

Our key idea can be derived as: \textit{the more real the fake face is, the more it can be defined as a difficult sample.} Specifically, as shown in yellow dotted frame, we defined two prototypes $P_{real}$ and $P_{fake}$ for real and fake features respectively and updated using EMA scheme defined as:

\begin{equation}
    P_{fake} = \alpha P_{fake} + (1-\alpha)k_{fake},
\end{equation}
\begin{equation}
    P_{real} = \alpha P_{real} + (1-\alpha)k_{real}.
\end{equation}

Then, we calculate the similarity of fake/fake feature and $P_{real}$/$P_{fake}$ as the basis to decide whether the feature enqueue. More formally, the screening progress can be derived as:

\begin{equation}
    \left\{
    \begin{aligned}
        &\delta(k_{fake},P_{real}) > \theta  , &  M_{f}\leftarrow k_{fake}\\
        &\delta(k_{real},P_{fake}) > \theta  , &  M_{r}\leftarrow k_{real},
    \end{aligned}
    \right.
    \label{equation:eq6}
\end{equation}

where $\theta$ is a threshold and the $\leftarrow$ represents the enqueue operation.
By doing this, the features in $M_f$ and $M_r$ of hard negatives have the following two properties: 1) their label is the same as the queue label. 2) it is hard for the model to distinguish their authenticity. 3) the quality of selected samples is higher. 
Thus, compared with ordinary feature queues without selection strategy, ours is more suitable for constructing true negative sample pairs.

\subsection{Intra-Instance Contrastive Learning.}
\label{section:4}
The aforementioned inter-instance contrastive learning module improves the generalization by contrastive among samples. To further promote generalized feature learning, we design an intra-instance contrastive learning module, which leverages the inconsistency of the forgery face by contrasting self-similarities within features.

Specifically, given forgery image $x_{fi}$, we first generate the pixel-level mask $m_i \in R^{H \times W}$ by subtracting its corresponding real image $x_{ri}^{'}$ : $m_i = |x_i - x_{i}^{'}|$. Then we resize the $m_i$ into the same spatial size as the feature map $f_q(v_1(x_{fi}))$ denoted as $m_i^{'} \in R^{H^{'} \times W^{'}}$. Subsequently, we segment the $f_q(v_1(x_{fi}))$ into real parts $P_r \in \{p_{r1},p_{r2},...p_{rn}\}$ and forgery parts $P_f \in \{p_{f1},p_{f2},...,p_{fk}\}$ using $m_i^{'}$, where $p_{f}, p_{r} \in {R^{C}}$ and $n,k$ denote the number of real and fake parts thus $n+k = H^{'}W^{'}$. 
Then the intra-instance contrastive loss for forgery features $L^{f}_{intra}$ is calculated based upon InfoNCE  as follows:

\begin{equation}
    L^{f}_{intra} = -\log\frac{\sum\limits_{i,j=1}^{n} e^{\delta(p_{ri},p_{rj}) / \tau}}{\sum\limits_{i,j}^{n} e^{\delta(p_{ri},p_{rj}) / \tau} + \sum\limits_{i=1}^{n}\sum\limits_{j=1}^{k} e^{\delta(p_{fi},p_{rj})/ \tau}
    },
    \label{equation:eq7}
\end{equation}

For real image $x_{ri}$, since all the features belongs to real, we expect for the self-similarity of $f_q(v_1(x_{ri}))$ become homogeneous. Thus, the intra-instance contrastive loss for real features can be obtained by:
\begin{equation}
    L^{r}_{intra} = -\log{sum(e^{f_q(v_1(x_{ri})) \odot f_q(v_1(x_{ri}))^{'T} / \tau })},
    \label{equation:eq8}
\end{equation}

where $\odot$ represents the gram matrices multiplication and $sum(.)$ represents the element-wise addition.$T$ denote the transpose operation.
The overall intra-instance contrastive loss $L_{intra}$ within a batch can be derived as :

\begin{equation}
    L_{intra} = L^{r}_{intra} + L^{f}_{intra},
    \label{equation:eq9}
\end{equation}

Different from ~\cite{chen2021local} which directly use similarity pattern as features, our proposed intra-instance contrastive loss enhances the inconsistency of forgery face by pulling away the similarity of real and fake parts and depress the influence of forgery irrelevant information via pulling the real pairs closer. Note that $L_{intra}$ do not aggregate the fake part together because we want to preserve the diversity of counterfeit traces.

\begin{table*}[!h]
    \renewcommand\arraystretch{1.1}
    \centering
    \resizebox{0.9\textwidth}{!}{
    \begin{tabular}{c|cc|cccccccc}
    \hline
    \multirow{2}*{Method} & \multicolumn{2}{c|}{\textit{FF++}}&\multicolumn{2}{c}{DFD} & \multicolumn{2}{c}{DFDC} & \multicolumn{2}{c}{Wild Deepfake} &\multicolumn{2}{c}{Celeb-DF}\\
    \cmidrule{2-3}
    \cmidrule{3-11}

    & AUC& EER &AUC& EER& AUC& EER& AUC& EER& AUC& EER\\
    \hline
    Xception& 99.09&3.77 &87.86& 21.04& 69.80& 35.41&66.17 &40.14 & 65.27& 38.77\\
    EN-b4   &99.22 &3.36& 87.37      & 21.99      & 70.12       & 34.54      & 61.04           & 45.34           & 68.52         & 35.61        \\
    Face X-ray& 87.40&-& 85.60&-& 70.00&      -      &       -          &         -        & 74.20          &       -       \\
    MLDG          &98.99 &3.46    & 88.14      & 21.34      & 71.86       & 34.44      & 64.12           & 43.27           & 74.56         & 30.81        \\
    
    F3-Net          & 98.10& 3.58 & 86.10     & 26.17       &     72.88       &      33.38      & 67.71          &      40.17           & 71.21        &     34.03         \\
    MAT(EN-b4) & 99.27&3.35& 87.58      & 21.73      & 67.34       & 38.31      & 70.15           & 36.53           & 76.65         & 32.83        \\
    GFF             & 98.36& 3.85 & 85.51      & 25.64      & 71.58       & 34.77      & 66.51           & 41.52            & 75.31         & 32.48        \\
    LTW            & 99.17& 3.32  & 88.56      & 20.57      & 74.58       & 33.81      & 67.12           & 39.22           & 77.14         & 29.34         \\
    
    Local-relation    & \textbf{99.46}&\textbf{3.01}& 89.24      & 20.32      & 76.53       & 32.41      & 68.76           & 37.50              & 78.26         & 29.67        \\
    \hline
    Ours              & 99.30&3.26& \textbf{91.66}      & \textbf{16.63}      & \textbf{76.71}       & \textbf{31.97}      & \textbf{71.14}           & \textbf{36.17}           & \textbf{82.30}          & \textbf{26.53}       \\
    \hline
    \end{tabular}
    }
    \caption{Cross-database evaluation from FF++(HQ) to DFD, DFDC, Wild Deepfake and Celeb-DF in terms of AUC and EER. The FF++ belongs to the intra-domain results while others represent to the unseen-domain.
    }
    \label{table:1}
    \end{table*}

\subsection{Overall Loss Function}
\label{section:5}
Considering both the cross-entropy loss based supervised learning branch and two InfoNCE losses based contrastive learning branches, the overall loss for our proposed method is:
\begin{equation}
    L_{all} = \phi (L_{inter} +L_{intra}) + (1-\phi) L_{ce},
    \label{equation:eq10}
\end{equation}

where $\phi$ is the hyper-parameters used to balance the cross-entropy loss and contrastive loss.

\section{Experiments}
\newcommand{\tabincell}[2]{\begin{tabular}{@{}#1@{}}#2\end{tabular}}

\begin{table}[t!]
    
    \renewcommand\arraystretch{1.1}
    \centering.
    \resizebox{0.9\columnwidth}{!}{
    \begin{tabular}{c|cc}
        \hline
        Method&\quad   FF++ \quad  \quad&Celeb-DF\quad\\
        \hline
        Meso-4&84.70&54.80\\
        Mesoinception4&83.00&53.60\\
        FWA&80.10&56.90\\
        Xception&95.50&65.50\\
        Multi-task&76.30&54.30\\
        SMIL&96.80&56.30\\
        Two Branch&93.18&	73.41\\
        EN-b4&96.39	&71.10\\
        Multi-Attention&96.41&72.50	\\
        GFF&95.73&74.12\\
        SPSL&96.91&	76.88\\
        \hline
        Ours&\textbf{96.97}&\textbf{81.00}\\
        \hline
    \end{tabular}
    }
     
    \caption{Cross-dataset evaluation from FF++(LQ) to deepfake class of FF++ and Celeb-DF in terms of AUC.}
    \vspace{-15pt}
    
    \label{table:2}
\end{table}

\subsection{Experimental Setting}
\textbf{Datasets.}
To evaluate our method, we conduct experiments on five famous challenging datasets: 
\textbf{FaceForensics++}~\cite{rossler2019faceforensics++} is a large-scale forgery face dataset containing $720$ videos for training and $280$ videos for validation or testing. There are four different face synthesis approaches in FaceForensics++, including two deep leaning based methods (\emph{DeepFakes} and \emph{NeuralTextures}) and two graphics-based approaches (\emph{Face2Face} and \emph{FaceSwap}), which is suitable for conducting generalization experiments. The videos in FaceForensics++ have two kinds of video quality: high-quality (quantization parameter equal to 23) and low-quality (quantization parameter equal to 40). \textbf{Celeb-DF}~\cite{li2019celeb} is another widely-used dataset, which contains $590$ real videos and 5639 fake videos. Forgery videos are generated by face swap for each pair of the $59$ subjects. \textbf{DFDC}~\cite{dolhansky2020deepfake} is a large-scale deepfake datasets which contain $1133$ real videos and $4080$ fake videos with various manipulated methods. \textbf{DFD} is a Deepfake based dataset that has $363$ real videos and $3068$ fake videos. \textbf{Wild Deepfake}~\cite{zi2020wilddeepfake} is a recently released forgery face dataset contains $3805$ real face sequences and $3509$ fake face sequences. All the videos are obtained from the internet. Therefore, wild deepfake has a variety of synthesis methods and backgrounds, as well as character ids. We use DSFD~\cite{li2019dsfd} to extract faces for all datasets and randomly select 50 frames from each video for testing and training.

\begin{table}[!t]
    \centering
    \renewcommand\arraystretch{1.0}
    \resizebox{0.90\columnwidth}{!}{
    \begin{tabular}{c|c|cccc}
    \hline
    Train   & Method       & DF    & F2F   & FS    & NT    \\
    \hline
    \multirow{4}{*}{DF}  & EN-b4 & \textit{99.97} & 76.32 & 46.24  & 72.72 \\
    & MAT          & \textit{99.92} & 75.23 & 40.61 & 71.08 \\
    & GFF          & \textit{99.87}	&76.89	&47.21&	72.88 \\
    & Ours    & \textit{\textbf{99.98}} & \textbf{77.13} & \textbf{61.01} & \textbf{75.01} \\
    \hline
    \multirow{3}{*}{F2F} & EN-b4 & 84.52 & \textit{99.20}  & 58.14 & 63.71 \\
    
    & MAT          & 86.15&	\textit{99.13}&	60.14&	64.59 \\
    & GFF          & 89.23	&\textit{99.10}&	\textbf{61.30}&	64.77 \\
    & Ours    & \textbf{91.91}  & \textit{\textbf{99.21}} & 59.58 & \textbf{66.67} \\
    \hline
    \multirow{3}{*}{FS}  & EN-b4 & 69.25 & 67.69 & \textit{99.89} & 48.61 \\
    & MAT          & 64.13	&66.39&\textit{99.67}	&	50.10 \\
    & GFF          &70.21&	68.72 &	\textit{99.85}&	49.91 \\
    & Ours    & \textbf{74.80}  &   \textbf{69.75}&\textit{\textbf{99.90}}  & \textbf{52.60}  \\
    \hline
    \multirow{3}{*}{NT}  & EN-b4 & 85.99 & 48.86 & 73.05 & \textit{98.25} \\
    & MAT          &  87.23	&48.22&	75.33&	\textit{98.66}  \\
    & GFF          & 88.49	&49.81&	74.31&	\textit{98.77} \\
    & Ours    & \textbf{91.23} & \textbf{52.13} & \textbf{79.31} & \textit{\textbf{98.97}}\\
    \hline
                         
    \end{tabular}
    }
    \caption{Cross-manipulation evaluation in terms of AUC. Diagonal results indicate the intra-domain performance.
    }
    \label{table:3}

    \end{table}
\begin{table*}[!t]
    \centering
    \renewcommand\arraystretch{1.1}
    \resizebox{0.8\textwidth}{!}{
        \scalebox{1}{
    \begin{tabular}{c|cccccccc}
        \hline
        \multirow{2}*{Method}& \multicolumn{2}{|l}{GID-DF (HQ)} & \multicolumn{2}{l}{GID-DF (LQ)} & \multicolumn{2}{l}{GID-F2F (HQ)} & \multicolumn{2}{l}{GID-F2F (LQ)} \\
        \cline{2-9}
    & ACC            & AUC           & ACC            & AUC           & ACC            & AUC            & ACC            & AUC            \\
    \hline
    EfficientNet         & 82.40           & 91.11         & 67.60           & 75.30          & 63.32          & 80.1           & 61.41           & 67.40           \\
    Focalloss        & 81.33          & 90.31          & 67.47           & 74.95         & 60.80           & 79.80           & 61.00             & 67.21           \\
    ForensicTransfer & 72.01          & -             & 68.20           & -             & 64.50           & -              & 55.00             & -              \\
    Multi-task       & 70.30           & -             & 66.76           & -             & 58.74           & -              & 56.50           & -              \\
    MLDG             & 84.21          & 91.82          & 67.15          & 73.12            & 63.46           & 77.10           & 58.12          & 61.70           \\
    LTW              & 85.60           & 92.70          & 69.15          & 75.60          & 65.60           & 80.20           & 65.70           & 72.40           \\
    \hline
    Ours             & \textbf{87.70}           & \textbf{94.9}         & \textbf{75.90}          & \textbf{83.82}         & \textbf{68.40}          & \textbf{82.93}          & \textbf{67.85}          & \textbf{75.07}    \\
    \hline     
    \end{tabular}
    }
    }
    \caption{Performance on multi-source manipulation evaluation, the protocols and results are from \cite{sun2021domain}. GID-DF means traning on the other three manipulated methods of FF++ and test on deepfakes class. The same for the others.}
    \label{table:4}
    \end{table*}

\noindent\textbf{Implement details.}
We resize the input face into $299\times299$, and use Adam optimizer to train the framework, where the weight decay is equal to $1e-5$ with betas of $0.9$ and $0.999$. The learning rate is set to $0.001$ and the batchsize is set to 32.
The EfficientNet-b4 ~\cite{tan2019efficientnet} pretrained on the ImageNet~\cite{deng2009imagenet} is used as our encoders $f_q$ and $f_k$. The exponential hyper-parameter $\beta$ is set to $0.99$. The temperature parameter $\tau$ of E.q.\,\ref{equation:eq3} is set to $0.07$ and the query size $|M|$ is set to $30000$. In addition, we set $0.9$ and $0.5$ for prototypes updating parameter $\alpha$ and threshold $\theta$. For the balanced weight $\phi$, we set $\phi=0.1$ for the first 5 epochs as the warm-up period under the guidance of $l_{ce}$, then the $\phi$ is set to $0.5$.

\subsection{Quantitative Results}
\textbf{Cross-dataset evaluation.}
To demonstrate the generalization of DCL, we conduct extensive cross-dataset evaluations.
Specifically, the models are trained on the FF++(HQ) and evaluated on the DFD, DFDC, Wild Deepfake, and Celeb-DF, respectively.
We compare DCL with several recently state-of-the-art methods, including \textbf{Xception}~\cite{chollet2017xception}, \textbf{EfficientNet-b4} ~\cite{tan2019efficientnet}, \textbf{Face X-ray}~\cite{tan2019efficientnet}, \textbf{MLDG}~\cite{li2018learning}, \textbf{F3-Net}~\cite{qian2020thinking}, \textbf{LTW}~\cite{sun2021domain}, \textbf{MTA}~\cite{zhao2021multi}, \textbf{Local-relation}~\cite{chen2021local} and \textbf{GFF}~\cite{luo2021generalizing}.

The results in Tab.\,\ref{table:1} show that our proposed DCL can significantly outperform the baseline model by around $5\%$ on average and obtain the state-of-the-art performance compared with recently general face forgery detection methods, especially on Celeb-DF. Compared with other cross-entropy loss based methods, our two contrastive losses can both diverse the intra-category invariance and enhance the inconsistency of forgery face, thus the generalization can be improved significantly.
In addition, the intra-domain performance(test on FF++) is better than the baseline model and close to the local-relation (drop $0.16\%$),
while our DCL outperforms it
by over $2\%$ on average in terms of AUC in unseen domain scenarios.
\begin{table}[!t]
    \centering
    \renewcommand\arraystretch{1.2}
    \resizebox{\columnwidth}{!}{
    \begin{tabular}{cccc|c|c}
        \hline
        Inter&Views& Hard & Intra &Celeb-DF&DFD\\
        \hline
        &\checkmark&&&74.12&88.32\\
        \hline
        \checkmark&&&&76.81&88.03\\
        \hline
        \checkmark&\checkmark&&&79.34&89.24\\
        \hline
        \checkmark&&\checkmark&&78.84&89.89\\
        \hline
        \checkmark&\checkmark&\checkmark&&80.30&90.12\\
        \hline
        \checkmark&\checkmark&\checkmark&\checkmark&\textbf{82.30}&\textbf{91.66}\\
        \hline
    \end{tabular}
    }
    \vspace{-2pt}
    \caption{Ablation study on the influence of 
    different components. Specifically, ``Inter" means inter-instance contrastive learning module, ``views" represents our special designed data views generation strategy, ``hard" indicate the hard sample generation, and ``Intra" is short for the Intra-instance contrastive learning module.}
    \label{table:5}
\end{table}

To further demonstrate the robustness of our method, we also evaluate the generalization when testing on low-quality images. Concretely, follow the setting of \cite{masi2020two}, we train our model on FF++ (LQ) and test it on Deepfakes class and Celeb-DF. The quantitative results are shown in Tab.\,\ref{table:2}, we can observe that our method obtain state-of-the-art performance especially in a cross-database setting. The DCL outperforms by $5\%$ compared with the recent SPSL and GFF on Celeb-DF and gets slight improvement on the intra-domain setting.

\begin{figure}[!t]
    \begin{center}
    \includegraphics[width=1\linewidth]{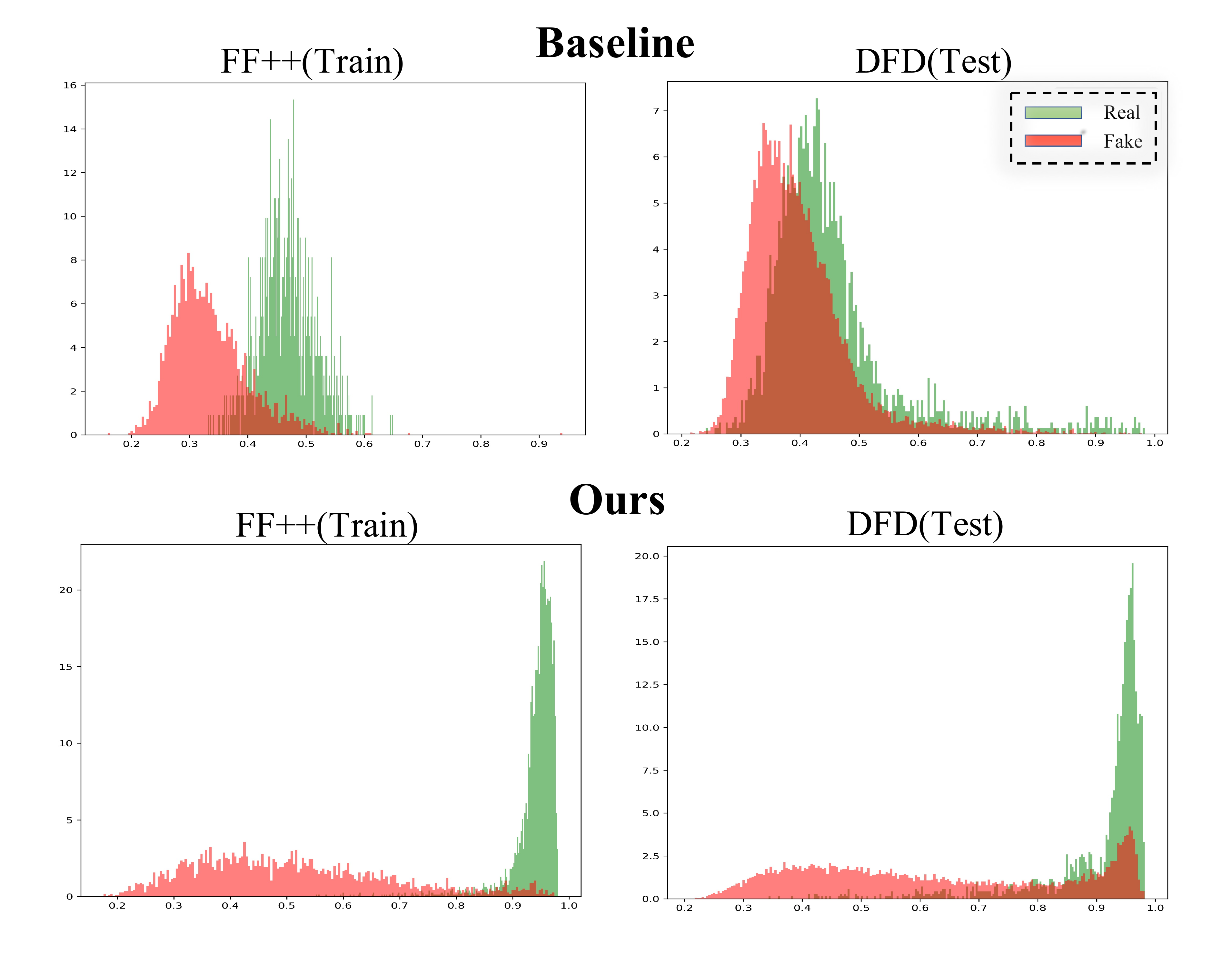}
    \end{center}
       \caption{
       Histogram of the average of self-similarity for intra-domain dataset (FF++) and unseen domain (DFD). The first row indicates the histogram of the baseline model (En-b4) while the second row represents that of our DCL.
       }
       \vspace{-6pt}
    \label{fig:hist}
\end{figure}

\noindent\textbf{Cross-manipulation evaluation.}
To further demonstrate the generalization among different manipulated methods, we conduct this experiment on the FF++(HQ) dataset. We train a model on one method of FF++ and test it on all four methods. For a fair comparison, we reimplement Multi-attentional and GFF with EfficientNet-b4 as backbones.

As shown in Tab.\,\ref{table:3}, our DCL outperforms the competitors in most cases, including both intra-manipulation (Diagonal of the table) results and cross-manipulation. Specifically, when the train on Deepfakes and test on Faceswap, our DCL achieves over $15\%$ performance gain on average in terms of AUC.
Since different manipulation leaves different traces of forgery, a generalization model should find essential differences between real and fake faces. Our method carefully designed this difference in inter-instance contrastive learning module and hard sample generation strategy, thus improving the cross-manipulation performance.

\noindent\textbf{Multi-source manipulation evaluation.}
In real-world applications, we usually have different manipulations samples to train while test on the unknown methods. We call these scenarios \textit{Multisource cross-manipulation}. To demonstrate the practicality of our method, we conduct experiments based on the benchmarks build by ~\cite{sun2021domain} and change our backbone to EfficientNet-b0 for a fair comparison. The results are reported in Tab.\,\ref{table:4}. Our DCL obtains state-of-the-art performance on all protocols in terms of AUC and ACC. In particular, our method outperforms by the recent LTW around $5\%$ on the low-quality version of FF++, which shows our method can guarantee generalization under different conditions and further demonstrates
the robustness of our framework.


\begin{figure}[t]
    \begin{center}
    \includegraphics[width=0.9\linewidth]{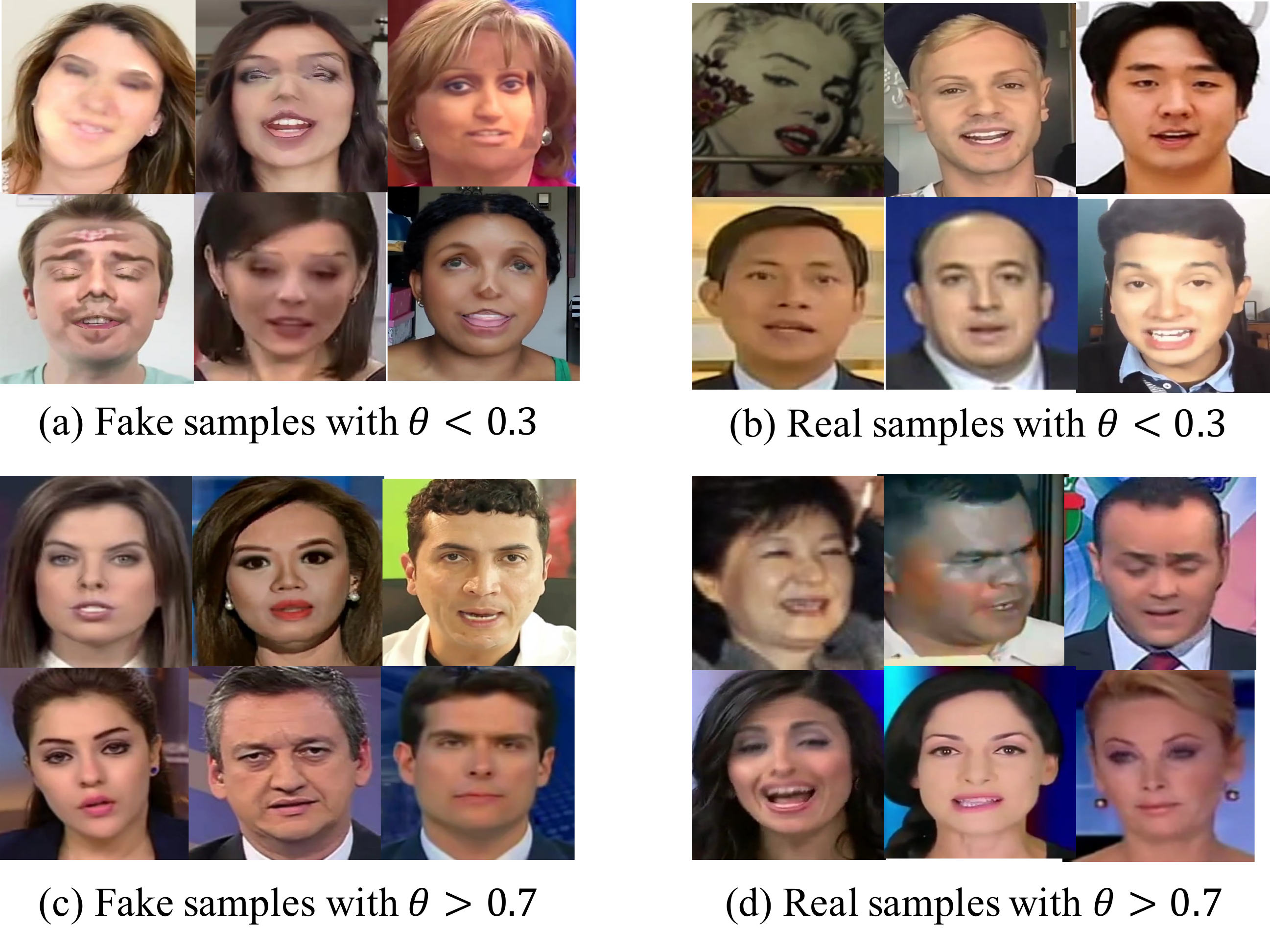}
    \end{center}
       \caption{
       Visualization of the hard sample strategy with low and high threshold.
       }
    \label{fig:hard}
\end{figure}
\subsection{Ablation Study}
To further explore the impact of different components of the DCL module, we split each part separately for verification. Specifically, we develop the following variants: 
1) baseline model with our specially designed data views generation.
baseline model with common data augmentation and Inter-ICL without hard sample selection scheme.
3) DCL without hard sample generation.
4) DCL without specially designed data views generation.
5) DCL without Inter-ICL.

The quantitative results on Celeb-DF and DFD are reported in Tab.\,\ref{table:5}, the metric is AUC.
The comparison between variant 2 and variant 3 can demonstrate the effectiveness of our specially designed data view transformation. 
The performance is further improved by $5\%$ on Celeb-DF dataset when using Inter-ICL, which shows the importance of feature distributions for generalization.
In addition, the hard sample selection scheme can bring around $1\%$ improvement on both two datasets. 
In particular, the performance boosts significantly after adding Intra-ICL, demonstrating the efficiency of mining the inconsistency within the instance.
Combining all the proposed components can achieve the best performance.

\begin{figure}[!t]
    \begin{center}
    \includegraphics[width=1\linewidth]{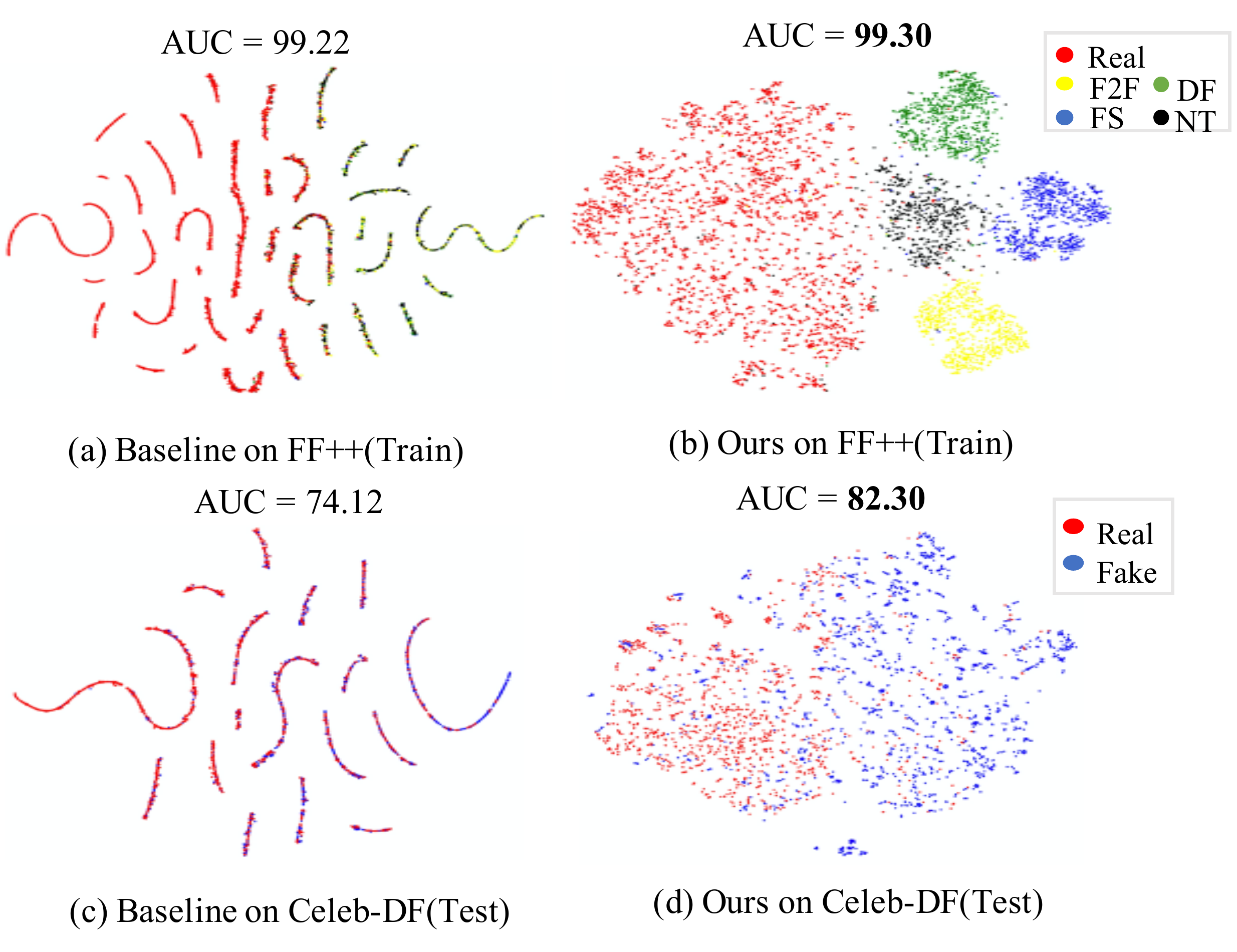}
    \end{center}
       \caption{ Feature distribution of baseline model (En-b4) and DCL on the  intra-domain dataset (FF++) and unseen domain dataset (Celeb-DF) via t-SNE. 
       }
    \label{fig:tsne}
    \vspace{-10pt}
\end{figure}
\subsection{Visualization}


\textbf{Visualization of self-similarity.}
As shown in Fig.\,\ref{fig:hist}, we draw the histogram of the real and fake self-similarity on the baseline model and DCL
to prove the effectiveness of our Intra-ICL. 
Specifically, we first train the DCL on FF++ (HQ), then the summation of self-similarity maps calculated by gram matrices is counted during the inference period. We can observe that the self-similarity of the baseline model lacks discriminative without additional constraints, while ours can be significantly separated in both intra-domain and unseen domain scenarios because of the Intra-ICL module. 

\noindent\textbf{Visualization of hard sample.}
Fig.\,\ref{fig:hard} represents the fake and real samples filtered by our hard sample selection strategy. We contrast the feature with the prototypes which is orthogonal to the original class and use threshold $\theta$ to select samples.
We can observe that the selection of fake samples ($\theta > 0.7$) are relatively high-quality compared with low threshold samples, which cannot be easily distinguished. These samples always contain more essential forgery clues which are commonly presented under forgery faces. For real faces, the selected samples usually have abnormal expressions or heavy makeup which is easy to confuse with the fake face.
And our framework takes these ``hard" samples as negative pairs to promote mining the essential forgery clues.

\noindent \textbf{Visualization of feature distribution.}
Our Inter-Contrastive learning aims to preserve the variations
by avoiding clustering of same class features. To verify this phenomenon, we draw feature distribution of cross-entropy and DCL based model using t-SNE~\cite{van2008visualizing} technique. The visualization results are shown in Fig.\,\ref{fig:tsne}. 
We can observe that the feature distributions of the same class are more dispersion than the baseline model and different manipulated methods are more separable, which improve the AUC of the unseen domain from $74.12\%$ to $82.30\%$.
This experiment demonstrates that the traditional supervised loss weakens the transferability due to the intra-category invariance, while our proposed DCL can improve the generalization by diverging the feature distribution.

\section{Conclusion}
    In this work, we propose a holistic learning framework, named Dual Contrastive Learning (DCL) for general face forgery detection. Specifically,
    we generate data views via specially designed data transformations as positive pairs and propose Inter-instance Contrastive Learning (Inter-ICL) and Intra-instance Contrastive Learning (Intra-ICL) to mine the association among instances and inconsistency within each sample. In addition, we also introduce the hard sample selection strategy to select informative hard negative samples and brings further advantage to DCL.
    Experiments on three settings demonstrate the significant superiority of our method over state-of-the-art methods.

\section*{Acknowledgments}
This work is supported by the National Science Fund for Distinguished Young Scholars (No.62025603), the National Natural Science Foundation of China (No.U1705262, No. 62072386, No. 62072387, No. 62072389, No. 62002305,  
No.61772443, No.61802324 and No.61702136), Guangdong Basic and Applied Basic Research Foundation(No.2019B1515120049), the Natural Science Foundation of Fujian Province of China (No.2021J01002), and the Fundamental Research Funds for the central universities (No. 20720200077, No. 20720200090 and No. 20720200091).

\bibliography{egbib}

\begin{thebibliography}{38}
\providecommand{\natexlab}[1]{#1}

\bibitem[{Afchar et~al.(2018)Afchar, Nozick, Yamagishi, and
  Echizen}]{afchar2018mesonet}
Afchar, D.; Nozick, V.; Yamagishi, J.; and Echizen, I. 2018.
\newblock Mesonet: a compact facial video forgery detection network.
\newblock In \emph{WIFS}, 1--7. IEEE.

\bibitem[{Bukchin et~al.(2021)Bukchin, Schwartz, Saenko, Shahar, Feris, Giryes,
  and Karlinsky}]{bukchin2021fine}
Bukchin, G.; Schwartz, E.; Saenko, K.; Shahar, O.; Feris, R.; Giryes, R.; and
  Karlinsky, L. 2021.
\newblock Fine-grained Angular Contrastive Learning with Coarse Labels.
\newblock In \emph{CVPR}, 8730--8740.

\bibitem[{Chen et~al.(2021)Chen, Yao, Chen, Ding, Li, and Ji}]{chen2021local}
Chen, S.; Yao, T.; Chen, Y.; Ding, S.; Li, J.; and Ji, R. 2021.
\newblock Local Relation Learning for Face Forgery Detection.
\newblock \emph{AAAI}.

\bibitem[{Chollet(2017)}]{chollet2017xception}
Chollet, F. 2017.
\newblock Xception: Deep learning with depthwise separable convolutions.
\newblock In \emph{CVPR}, 1251--1258.

\bibitem[{Deng et~al.(2009)Deng, Dong, Socher, Li, Li, and
  Fei-Fei}]{deng2009imagenet}
Deng, J.; Dong, W.; Socher, R.; Li, L.-J.; Li, K.; and Fei-Fei, L. 2009.
\newblock Imagenet: A large-scale hierarchical image database.
\newblock In \emph{CVPR}, 248--255. Ieee.

\bibitem[{Dolhansky et~al.(2020)Dolhansky, Bitton, Pflaum, Lu, Howes, Wang, and
  Ferrer}]{dolhansky2020deepfake}
Dolhansky, B.; Bitton, J.; Pflaum, B.; Lu, J.; Howes, R.; Wang, M.; and Ferrer,
  C.~C. 2020.
\newblock The DeepFake Detection Challenge Dataset.
\newblock \emph{arXiv preprint arXiv:2006.07397}.

\bibitem[{Frank et~al.(2020)Frank, Eisenhofer, Sch{\"o}nherr, Fischer, Kolossa,
  and Holz}]{frank2020leveraging}
Frank, J.; Eisenhofer, T.; Sch{\"o}nherr, L.; Fischer, A.; Kolossa, D.; and
  Holz, T. 2020.
\newblock Leveraging frequency analysis for deep fake image recognition.
\newblock In \emph{International Conference on Machine Learning}, 3247--3258.
  PMLR.

\bibitem[{Fridrich and Kodovsky(2012)}]{fridrich2012rich}
Fridrich, J.; and Kodovsky, J. 2012.
\newblock Rich models for steganalysis of digital images.
\newblock \emph{IEEE Transactions on Information Forensics and Security}, 7(3):
  868--882.

\bibitem[{Gu et~al.(2021)Gu, Chen, Yao, Ding, Li, Huang, and
  Ma}]{gu2021spatiotemporal}
Gu, Z.; Chen, Y.; Yao, T.; Ding, S.; Li, J.; Huang, F.; and Ma, L. 2021.
\newblock Spatiotemporal Inconsistency Learning for DeepFake Video Detection.
\newblock In \emph{Proceedings of the 29th ACM International Conference on
  Multimedia}, 3473--3481.

\bibitem[{Gutmann and Hyv{\"a}rinen(2010)}]{gutmann2010noise}
Gutmann, M.; and Hyv{\"a}rinen, A. 2010.
\newblock Noise-contrastive estimation: A new estimation principle for
  unnormalized statistical models.
\newblock In \emph{Proceedings of the thirteenth international conference on
  artificial intelligence and statistics}, 297--304. JMLR Workshop and
  Conference Proceedings.

\bibitem[{He et~al.(2020)He, Fan, Wu, Xie, and Girshick}]{he2020momentum}
He, K.; Fan, H.; Wu, Y.; Xie, S.; and Girshick, R. 2020.
\newblock Momentum contrast for unsupervised visual representation learning.
\newblock In \emph{CVPR}, 9729--9738.

\bibitem[{Khosla et~al.(2020)Khosla, Teterwak, Wang, Sarna, Tian, Isola,
  Maschinot, Liu, and Krishnan}]{khosla2020supervised}
Khosla, P.; Teterwak, P.; Wang, C.; Sarna, A.; Tian, Y.; Isola, P.; Maschinot,
  A.; Liu, C.; and Krishnan, D. 2020.
\newblock Supervised contrastive learning.
\newblock \emph{arXiv preprint arXiv:2004.11362}.

\bibitem[{Li et~al.(2018)Li, Yang, Song, and Hospedales}]{li2018learning}
Li, D.; Yang, Y.; Song, Y.-Z.; and Hospedales, T.~M. 2018.
\newblock Learning to generalize: Meta-learning for domain generalization.
\newblock In \emph{AAAI}.

\bibitem[{Li et~al.(2019{\natexlab{a}})Li, Wang, Wang, Tai, Qian, Yang, Wang,
  Li, and Huang}]{li2019dsfd}
Li, J.; Wang, Y.; Wang, C.; Tai, Y.; Qian, J.; Yang, J.; Wang, C.; Li, J.; and
  Huang, F. 2019{\natexlab{a}}.
\newblock DSFD: dual shot face detector.
\newblock In \emph{CVPR}, 5060--5069.

\bibitem[{Li et~al.(2020)Li, Bao, Zhang, Yang, Chen, Wen, and Guo}]{li2020face}
Li, L.; Bao, J.; Zhang, T.; Yang, H.; Chen, D.; Wen, F.; and Guo, B. 2020.
\newblock Face x-ray for more general face forgery detection.
\newblock In \emph{CVPR}, 5001--5010.

\bibitem[{Li, Chang, and Lyu(2018)}]{Li2018InIO}
Li, Y.; Chang, M.-C.; and Lyu, S. 2018.
\newblock In Ictu Oculi: Exposing AI Created Fake Videos by Detecting Eye
  Blinking.
\newblock \emph{2018 IEEE International Workshop on Information Forensics and
  Security (WIFS)}, 1--7.

\bibitem[{Li et~al.(2019{\natexlab{b}})Li, Yang, Sun, Qi, and
  Lyu}]{li2019celeb}
Li, Y.; Yang, X.; Sun, P.; Qi, H.; and Lyu, S. 2019{\natexlab{b}}.
\newblock Celeb-df: A new dataset for deepfake forensics.
\newblock \emph{arXiv preprint arXiv:1909.12962}.

\bibitem[{Liu et~al.(2021{\natexlab{a}})Liu, Li, Zhou, Chen, He, Xue, Zhang,
  and Yu}]{liu2021spatial}
Liu, H.; Li, X.; Zhou, W.; Chen, Y.; He, Y.; Xue, H.; Zhang, W.; and Yu, N.
  2021{\natexlab{a}}.
\newblock Spatial-phase shallow learning: rethinking face forgery detection in
  frequency domain.
\newblock In \emph{CVPR}, 772--781.

\bibitem[{Liu et~al.(2021{\natexlab{b}})Liu, Zhang, Yao, Sheng, Ding, Tai, Li,
  Xie, and Ma}]{liu2021dual}
Liu, S.; Zhang, K.-Y.; Yao, T.; Sheng, K.; Ding, S.; Tai, Y.; Li, J.; Xie, Y.;
  and Ma, L. 2021{\natexlab{b}}.
\newblock Dual reweighting domain generalization for face presentation attack
  detection.
\newblock \emph{IJCAI}.

\bibitem[{Lo et~al.(2021)Lo, Chang, Chiu, Huang, Chen, Chang, and
  Jou}]{lo2021clcc}
Lo, Y.-C.; Chang, C.-C.; Chiu, H.-C.; Huang, Y.-H.; Chen, C.-P.; Chang, Y.-L.;
  and Jou, K. 2021.
\newblock CLCC: Contrastive Learning for Color Constancy.
\newblock In \emph{CVPR}, 8053--8063.

\bibitem[{Luo et~al.(2021)Luo, Zhang, Yan, and Liu}]{luo2021generalizing}
Luo, Y.; Zhang, Y.; Yan, J.; and Liu, W. 2021.
\newblock Generalizing Face Forgery Detection with High-frequency Features.
\newblock In \emph{CVPR}, 16317--16326.

\bibitem[{Masi et~al.(2020)Masi, Killekar, Mascarenhas, Gurudatt, and
  AbdAlmageed}]{masi2020two}
Masi, I.; Killekar, A.; Mascarenhas, R.~M.; Gurudatt, S.~P.; and AbdAlmageed,
  W. 2020.
\newblock Two-branch recurrent network for isolating deepfakes in videos.
\newblock In \emph{ECCV}, 667--684. Springer.

\bibitem[{Matern, Riess, and Stamminger(2019)}]{matern2019exploiting}
Matern, F.; Riess, C.; and Stamminger, M. 2019.
\newblock Exploiting visual artifacts to expose deepfakes and face
  manipulations.
\newblock In \emph{WACVW}, 83--92. IEEE.

\bibitem[{Qian et~al.(2020)Qian, Yin, Sheng, Chen, and Shao}]{qian2020thinking}
Qian, Y.; Yin, G.; Sheng, L.; Chen, Z.; and Shao, J. 2020.
\newblock Thinking in Frequency: Face Forgery Detection by Mining
  Frequency-aware Clues.
\newblock In \emph{ECCV}, 86--103. Springer.

\bibitem[{Robinson et~al.(2021)Robinson, Chuang, Sra, and
  Jegelka}]{robinson2020contrastive}
Robinson, J.; Chuang, C.-Y.; Sra, S.; and Jegelka, S. 2021.
\newblock Contrastive learning with hard negative samples.
\newblock \emph{ICLR}.

\bibitem[{Rossler et~al.(2019)Rossler, Cozzolino, Verdoliva, Riess, Thies, and
  Nie{\ss}ner}]{rossler2019faceforensics++}
Rossler, A.; Cozzolino, D.; Verdoliva, L.; Riess, C.; Thies, J.; and
  Nie{\ss}ner, M. 2019.
\newblock Faceforensics++: Learning to detect manipulated facial images.
\newblock In \emph{ICCV}, 1--11.

\bibitem[{Stehouwer et~al.(2019)Stehouwer, Dang, Liu, Liu, and
  Jain}]{stehouwer2019detection}
Stehouwer, J.; Dang, H.; Liu, F.; Liu, X.; and Jain, A. 2019.
\newblock On the detection of digital face manipulation.
\newblock \emph{arXiv preprint arXiv:1910.01717}.

\bibitem[{Sun et~al.(2021)Sun, Liu, Ye, Liu, Gao, Shao, and Ji}]{sun2021domain}
Sun, K.; Liu, H.; Ye, Q.; Liu, J.; Gao, Y.; Shao, L.; and Ji, R. 2021.
\newblock Domain General Face Forgery Detection by Learning to Weight.
\newblock In \emph{AAAI}, volume~35, 2638--2646.

\bibitem[{Tan and Le(2019)}]{tan2019efficientnet}
Tan, M.; and Le, Q.~V. 2019.
\newblock Efficientnet: Rethinking model scaling for convolutional neural
  networks.
\newblock \emph{ICML}.

\bibitem[{Thies et~al.(2015)Thies, Zollh{\"o}fer, Nie{\ss}ner, Valgaerts,
  Stamminger, and Theobalt}]{thies2015real}
Thies, J.; Zollh{\"o}fer, M.; Nie{\ss}ner, M.; Valgaerts, L.; Stamminger, M.;
  and Theobalt, C. 2015.
\newblock Real-time expression transfer for facial reenactment.
\newblock \emph{ACM Trans. Graph.}, 34(6): 183--1.

\bibitem[{Van~der Maaten and Hinton(2008)}]{van2008visualizing}
Van~der Maaten, L.; and Hinton, G. 2008.
\newblock Visualizing data using t-SNE.
\newblock \emph{Journal of machine learning research}, 9(11).

\bibitem[{Wang and Isola(2020)}]{wang2020understanding}
Wang, T.; and Isola, P. 2020.
\newblock Understanding contrastive representation learning through alignment
  and uniformity on the hypersphere.
\newblock In \emph{ICML}, 9929--9939. PMLR.

\bibitem[{Wang et~al.(2021)Wang, Zhou, Yu, Dai, Konukoglu, and
  Van~Gool}]{wang2021exploring}
Wang, W.; Zhou, T.; Yu, F.; Dai, J.; Konukoglu, E.; and Van~Gool, L. 2021.
\newblock Exploring cross-image pixel contrast for semantic segmentation.
\newblock \emph{arXiv preprint arXiv:2101.11939}.

\bibitem[{Wang et~al.(2020)Wang, Yao, Ding, and Ma}]{wang2020face}
Wang, X.; Yao, T.; Ding, S.; and Ma, L. 2020.
\newblock Face manipulation detection via auxiliary supervision.
\newblock In \emph{International Conference on Neural Information Processing},
  313--324. Springer.

\bibitem[{Yang, Li, and Lyu(2019)}]{yang2019exposing}
Yang, X.; Li, Y.; and Lyu, S. 2019.
\newblock Exposing deep fakes using inconsistent head poses.
\newblock In \emph{ICASSP}, 8261--8265. IEEE.

\bibitem[{Zhao et~al.(2021)Zhao, Zhou, Chen, Wei, Zhang, and
  Yu}]{zhao2021multi}
Zhao, H.; Zhou, W.; Chen, D.; Wei, T.; Zhang, W.; and Yu, N. 2021.
\newblock Multi-attentional Deepfake Detection.
\newblock \emph{CVPR}.

\bibitem[{Zhao et~al.(2020)Zhao, Wu, Lau, and Lin}]{zhao2020makes}
Zhao, N.; Wu, Z.; Lau, R.~W.; and Lin, S. 2020.
\newblock What makes instance discrimination good for transfer learning?
\newblock \emph{ICLR}.

\bibitem[{Zi et~al.(2020)Zi, Chang, Chen, Ma, and Jiang}]{zi2020wilddeepfake}
Zi, B.; Chang, M.; Chen, J.; Ma, X.; and Jiang, Y.-G. 2020.
\newblock Wilddeepfake: A challenging real-world dataset for deepfake
  detection.
\newblock In \emph{ACM MM}, 2382--2390.

\end{thebibliography}

\bigskip

\end{document}